\newif\ifarxiv
\arxivtrue   

\ifarxiv
  \documentclass[sigconf,nonacm]{acmart}
\else
  \documentclass[sigconf]{acmart}
\fi

\AtBeginDocument{%
  }

\ifarxiv
  \renewcommand\footnotetextcopyrightpermission[1]{}
\else
  \setcopyright{none}
  \acmConference[ACM CAIS 2026]{ACM Conference on AI and Agentic Systems}{May 26--29, 2026}{San Jose, CA}
\fi

\usepackage{xspace}

\usepackage{subcaption}
\usepackage{graphicx}

\usepackage{listings}
\usepackage{tikz}
\usetikzlibrary{positioning}
\usepackage{tcolorbox}
\usepackage{multicol}
\usepackage{xcolor}

\ifarxiv
  \usepackage{listings}
  \lstnewenvironment{codeblock}{
    \lstset{
      basicstyle=\ttfamily\footnotesize,
      breaklines=true,
      columns=fullflexible
    }
  }{}
\else
  \usepackage{minted}
  \newenvironment{codeblock}
    {\VerbatimEnvironment\begin{minted}[
      fontsize=\footnotesize,
      numbersep=6pt,
      breaklines,
      style=tango,
      mathescape
    ]{go}}
    {\end{minted}}
\fi

\usepackage{enumitem}

\usepackage{xcolor}

\usepackage{marginnote}

\usepackage{float}
\usepackage{subcaption}

\newcommand{\comp}[1]{\vspace{2pt}\noindent\textbf{#1.} }

\newcommand{\rex}{\fcolorbox{gray!60}{gray!15}{\textbf{Example.}}}

\definecolor{stagegreen}{RGB}{184, 242, 184}
\definecolor{stageyellow}{RGB}{255, 235, 170}
\definecolor{stagelime}{RGB}{210, 255, 170}
\definecolor{stagepink}{RGB}{242, 200, 200}
\definecolor{stageblue}{RGB}{200,220,245}

\newcommand{\hayder}[1]{\textcolor{magenta}{hayder: #1}}
\newcommand{\orla}{\textrm{Orla}\xspace}
\newcommand{\minlan}[1]{\textcolor{red}{Minlan: #1}}

\newtcolorbox{runningexample}[1][]{
  colback=black!1,
  colframe=black!35,
  boxrule=0.3pt,
  sharp corners,
  title={#1},
  fonttitle=\bfseries,
  coltitle=black,
  colbacktitle=black!8,
  left=3pt,
  right=3pt,
  top=2pt,
  bottom=2pt,
  boxsep=1pt
}

\newif\ifshortversion
\shortversiontrue

\newcommand{\fullonly}[1]{\ifshortversion\else#1\fi}
\newcommand{\shortonly}[1]{\ifshortversion#1\fi}

\begin{document}


\title{\includegraphics[height=1.2em]{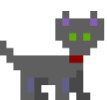}\ \orla: A Library for Serving LLM-Based Multi-Agent Systems}

\author{Rana Shahout}
\authornote{Both authors contributed equally to this research.}
\affiliation{%
 \institution{Harvard University}
 \state{MA}
 \country{USA}
}

\author{Hayder Tirmazi}
\authornotemark[1]
\affiliation{%
  \institution{Boston University}
\state{MA}
 \country{USA}
}

\author{Minlan Yu}
\affiliation{%
 \institution{Harvard University}
 \state{MA}
 \country{USA}
}

\author{Michael Mitzenmacher}
\affiliation{%
 \institution{Harvard University}
 \state{MA}
 \country{USA}
}

\begin{abstract}



We introduce \orla, a library for constructing and running LLM-based agentic systems. 
Modern agentic applications consist of workflows that combine multiple LLM inference steps, tool calls, and heterogeneous infrastructure. 
Today, developers typically build these systems by manually composing orchestration code with LLM serving engines and tool execution logic.

\orla provides a general abstraction that separates request execution from workflow-level policy. 
It acts as a serving layer above existing LLM inference engines: developers define workflows composed of stages, while \orla manages how those stages are mapped, executed, and coordinated across models and backends. 
It provides agent-level control through three mechanisms: a \emph{stage mapper}, which assigns each stage to an appropriate model and backend; a \emph{workflow orchestrator}, which schedules stages and manages their resources and context; and a \emph{memory manager}, which manages inference state such as the KV cache across workflow boundaries.

We demonstrate \orla with a customer support workflow that exercises many of its capabilities. We evaluate \orla on two datasets, showing that stage mapping improves latency and cost compared to a single-model vLLM baseline, while workflow-level cache management reduces time-to-first-token.


\end{abstract}


\maketitle

\ifarxiv
  \pagestyle{plain}
\fi

\section{Introduction}
Agentic systems built on large language models (LLMs) have demonstrated strong performance across diverse applications, including code generation~\cite{jimenez2023swe,xia2025live}, finance~\cite{wu2023bloomberggpt,Bloomberg2026ASKB}, and scientific discovery~\cite{gottweis2025towards}. They are also widely viewed as a promising path toward artificial general intelligence (AGI)~\cite{gao2025survey}.
These systems solve complex tasks by executing \emph{workflows}: iterative cycles of reasoning, planning, and acting, where the system consumes context, generates intermediate outputs, invokes external tools, and incorporates tool results into subsequent inference steps.  We refer to each step in this workflow as a \emph{stage}. As tasks grow in complexity, the field has increasingly turned to multi-agent systems, where specialized agents, potentially backed by different models, collaborate by exchanging context and intermediate results across workflow stages.

Despite rapid adoption, the current infrastructure for serving agentic systems remains poorly suited to their demands. Today's agentic inference stacks are loosely assembled from isolated components, pairing application-level orchestration code with LLM inference engines~\cite{shahout2024don, shahout2024fast, mitzenmacher2025queueing} such as vLLM~\cite{kwon2023efficient} or SGLang~\cite{kang2026thunderagent}. These serving systems optimize request-level execution through prefill and decode scheduling~\cite{patel2024splitwise}, continuous batching~\cite{yu2022orca}, and KV cache management to improve throughput and reduce latency. These optimizations are necessary, but they address only a subset of the system challenges that arise in agentic workloads.


Serving agentic systems must address two closely related challenges: \textbf{heterogeneity} across workflow stages and \textbf{coordination and decision-making} across the components that execute them.
Heterogeneity appears along several dimensions, including the models used, the infrastructure on which stages execute, and the characteristics of the workloads themselves. Within a single workflow, different stages often have different computational requirements. For example, a routing or summarization step may benefit from a small, fast model, whereas a synthesis or code-generation step requires a larger, more capable one. At the same time, stages may execute on different infrastructure. Modern deployments frequently span multiple serving backends and hardware environments, such as SGLang, vLLM, or Ollama. Finally, agentic workloads vary in task complexity, input characteristics, and performance objectives.

Stage decomposition further leads to the need for coordination and decision-making.  
Agentic systems must decide which models to use, where to execute each stage, and how to schedule and manage resources across the workflow. While existing request-level serving systems can optimize the execution of individual inference calls, they lack mechanisms to coordinate decisions across stages, models, and infrastructure within a workflow.

Motivated by the need for a new serving system that operates at the workflow level to coordinate decisions across stages, we introduce \orla, a serving system for agentic workflows that sits above existing LLM serving backends. Our key insight is to separate request execution from agent-level policy through three components:

\textbf{Stage mapper.}
The stage mapper 
aims to assign each stage to the most appropriate model and backend based on its computational requirements, performance, and available infrastructure. \orla supports both explicit developer-specified mappings and dynamic routing through pluggable mappers.

\textbf{Workflow orchestrator.}
The workflow orchestrator coordinates stage execution across workflows. It operates at two levels: scheduling and resource and context management. The scheduler determines the execution order of stages across workflows and enforces stage dependencies within each workflow. Resource and context management dispatch ready stages concurrently, keep backends utilized during tool-call waits, and reroute stages to alternative backends under load. As workflows progress, the orchestrator also manages evolving context across stages; iterative calls often share large context prefixes, while tool outputs and stage transitions may introduce new information that must be incorporated into subsequent stages.

\textbf{Memory manager.}
KV-cache decisions in agentic workloads should not be made at the level of individual inference calls. Consecutive stages in a workflow may reuse the same context, while multiple workflows must share limited GPU memory. This creates a tradeoff between preserving KV cache for reuse, which avoids redundant prefill and reduces latency, and reclaiming memory, which frees GPU capacity for other workflows. The memory manager coordinates the life-cycle of inference state across workflow boundaries by deciding when to preserve KV cache for reuse across stages and when to reclaim memory under pressure. These decisions rely on workflow-level signals, such as stage dependencies, predicted reuse, backend transitions, and workflow completion, that are invisible to a request-level backend.


\emph{Together, these components form a lightweight library for constructing agentic systems. Developers define workflows using simple stage abstractions, while \orla handles the underlying coordination of models, infrastructure, and resources.} Our evaluation shows that stage mapping improves latency and cost, while workflow-level cache management reduces first-token latency.

\textbf{\orla as a library and research testbed.}
A key design decision in \orla is to treat all backends as plug-and-play modules accessed through a uniform interface. \orla communicates with backends via an OpenAI-compatible HTTP API, enabling it to interact uniformly with systems such as SGLang, Ollama, or lightweight simulators. Switching between backends requires only changing the endpoint URL in the stage configuration. No modifications are required to the workflow definition, scheduling policies, or memory manager. 
This makes \orla suitable both for production serving and as a research testbed. Researchers studying agentic scheduling, KV cache management, or multi-agent coordination can run experiments on commodity hardware without GPU infrastructure. A simulated backend that models prefill and decode latency parametrically isolates \orla's orchestration behavior from the variance of real LLM inference, enabling controlled and reproducible experiments. The live demo artifact exploits this property: on machines without a GPU, it runs the full customer support workflow using lightweight Ollama models on a CPU.


\fullonly{\section{Related Work}

\noindent\textbf{LLM serving systems.}
SGLang~\cite{zheng2024sglang} and vLLM~\cite{kwon2023efficient}
optimize request-level serving through KV cache management techniques such as RadixAttention and PagedAttention,
continuous batching, and structured decoding. \orla builds on
these systems as backends and adds workflow-level scheduling,
cross-backend routing, and agent-aware cache management that
operate above the request level.

\medskip\noindent\textbf{Agent orchestration frameworks.}
LangGraph~\cite{langgraph} and AutoGen~\cite{wu2024autogen} provide application-level
abstractions for composing multi-agent workflows. These
frameworks focus on defining agent roles, conversation
patterns, tool integrations, and operational concerns such as
persistence and fault tolerance. While they allow developers to assign different models to different agents, they treat the serving layer as a black box and do not make serving-level decisions such as dynamic request routing based on complexity,
two-level scheduling across shared backends, or KV cache
lifecycle management at workflow boundaries. \orla's contributions
are complementary: it operates below these frameworks and above
the inference engine, providing serving-layer optimizations
that orchestration frameworks cannot express.

The work closest to \orla is
ThunderAgent~\cite{kang2026thunderagent}, which abstracts
agentic workflows as stateful programs and introduces a
program-aware scheduler that mitigates KV cache thrashing
and memory imbalance across GPU nodes, achieving significant
throughput gains on vLLM and SGLang backends.
ThunderAgent models programs as flat sequences of reasoning
and acting phases and schedules a single model across nodes.
\orla addresses a more general formulation: it models
multi-stage workflows as DAGs, routes heterogeneous models
across diverse backends within a single workflow, and
applies per-stage scheduling policies. A flat,
single-model program is a special case of this execution
model. The two systems are also complementary. For example,
ThunderAgent can provide deep optimization of GPU memory
scheduling under high concurrency, while \orla can serve as a
cross-backend, multi-model workflow-level control plane.

\medskip\noindent\textbf{Compound AI systems.}
DSPy~\cite{khattab2023dspy} compiles declarative language model programs
into optimized prompts and can use SGLang as a backend.
\orla addresses a different layer: rather than optimizing
prompts, it optimizes serving decisions such as routing,
scheduling, and cache management for multi-stage agentic workflows.}
\section{\orla Design}\label{sec:orla}

\begin{figure}
  \centering
  \includegraphics[width=0.9\linewidth]{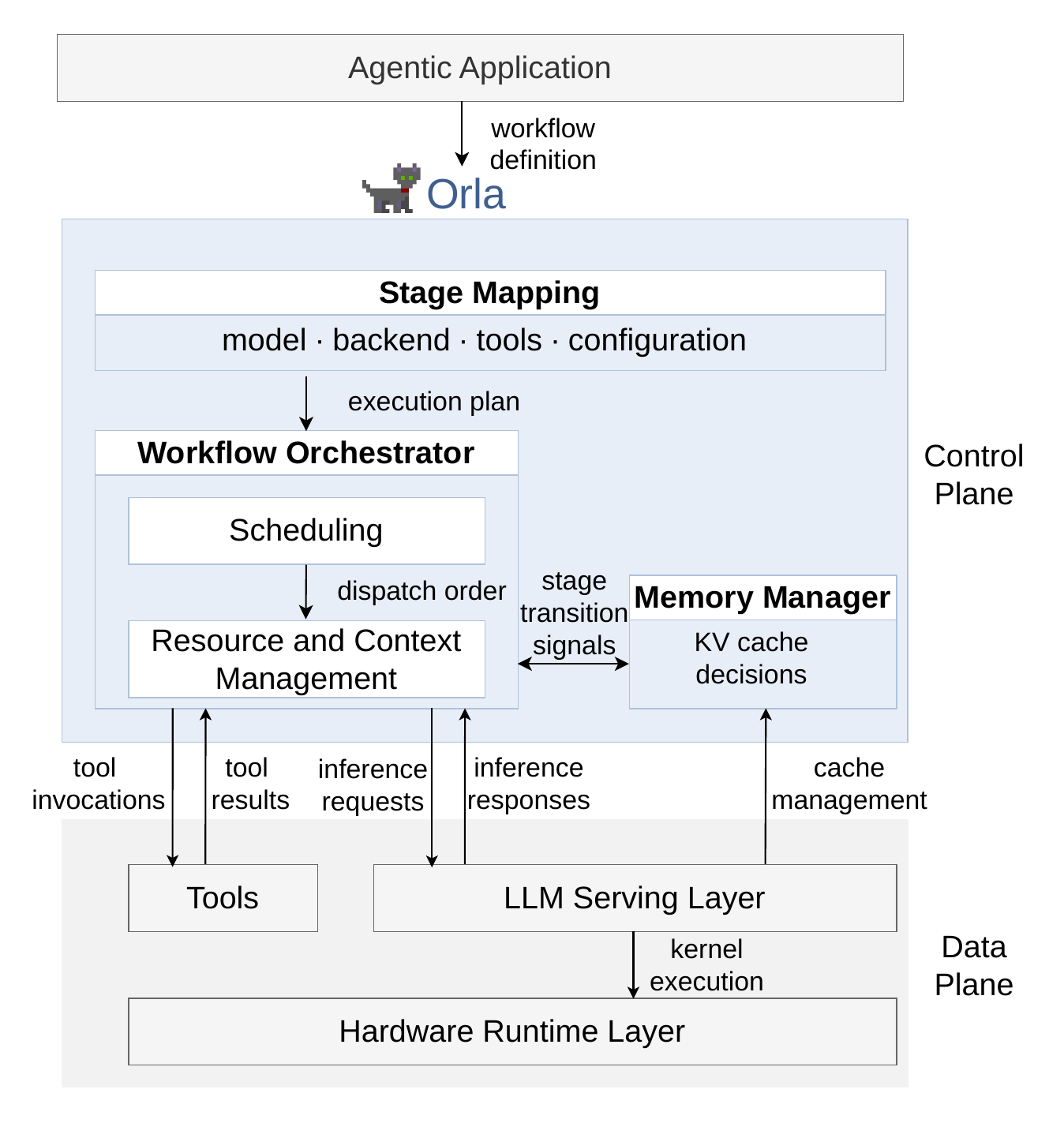}
  \caption{\orla design.}
  \label{fig:orla_overview}
\end{figure}

\subsection{Core Abstractions}

\fullonly{\emph{Stage.} A stage is the primary execution unit in an agentic workflow. It begins with a context (e.g., prompt and history), performs LLM inference, and may invoke external tools whose outputs are incorporated into subsequent inference steps. Execution therefore proceeds as a loop: the model generates an output, the system executes any tool invocations it contains, and the results are fed back into the context until the model signals completion or a turn limit is reached. In the simplest case, a stage completes in a single inference call without invoking tools.
Each stage encapsulates the configuration required to execute this loop, including (1) a target LLM backend, (2) inference parameters such as temperature, maximum tokens, and structured output schema, and (3) an optional set of compatible tools. Every stage is constructed with a name and a globally unique identifier, which is attached to its execution requests and enables per-stage queue management and scheduling.}

\shortonly{\emph{Stage.}
A stage is the basic execution unit in a workflow. It performs LLM inference on a given context and may iteratively invoke tools whose results are fed back into the model until completion. Each stage specifies its model backend, inference parameters, and optional tools.}

\fullonly{\noinden\emph{Workflow.}
A workflow defines the execution plan for an agentic task. It specifies the stages required to complete the task and the dependencies between them, determining how inference and tool execution progress from one stage to the next. In the current design, workflows are represented as directed acyclic graphs (DAGs) of stages, where edges encode execution dependencies. This abstraction captures common patterns such as sequential pipelines, parallel fan-out execution, and more complex dependency structures.
Developers construct workflows by registering stages and declaring their dependencies ahead of time. A stage becomes eligible for execution once all of its upstream dependencies have completed. Context flows between stages through builder functions, which construct a downstream stage’s prompt or message list from the outputs of its upstream stages. Dynamic behavior in \orla arises from runtime serving decisions—such as stage mapping, backend selection, and scheduling—rather than from modifying the workflow graph itself.}

\shortonly{\noindent\emph{Workflow.}
A workflow defines the execution plan for an agentic task. It is represented as a directed acyclic graph (DAG) of stages, where edges encode execution dependencies. A stage becomes eligible to run once all upstream stages complete, and its input context is constructed from their outputs.}

\fullonly{\noindent\emph{Backend.}
A backend is an LLM inference endpoint that executes stage requests. Backends may differ in model, hardware environment, and serving stack, such as SGLang on a datacenter GPU or Ollama on a local CPU. \orla treats backends as interchangeable modules accessed through a uniform interface, allowing stages in the same workflow to execute across heterogeneous infrastructure.}

\shortonly{\noindent\emph{Backend.}
A backend is an LLM inference endpoint that executes stage requests. Backends may differ in model, hardware, or serving stack (e.g., SGLang or Ollama), but \orla accesses them through a uniform interface, enabling workflows to run across heterogeneous infrastructure.}

\vspace{-6pt}

\subsection{System Overview}

\fullonly{We view the serving stack for agentic systems as four layers (Figure~\ref{fig:orla_overview}). The first layer is a hardware runtime responsible for executing the tensor programs that underlie LLM inference, such as CUDA for GPU inference, Metal for Apple platforms, or ZenDNN for CPU inference. The second layer is an LLM serving layer that processes individual generation requests on the hardware runtime, exemplified by SGLang~\cite{}, vLLM~\cite{}, and Ollama~\cite{}. The third layer is \orla, which is our contribution, that manages the execution of multi-agent workflows across one or more LLM serving backends, and the last layer is the application level.

\orla treats LLM serving backends as programmable execution engines and supplies agent-aware decisions about what to run, where to run it, and in what order. This shifts the optimization target from maximizing per-request throughput to minimizing end-to-end workflow completion time and cost.}

\orla contains three system components that operate on these abstractions: a \emph{stage mapper}, a \emph{workflow orchestrator} and a \emph{memory manager}.
\shortonly{(Figure~\ref{fig:orla_overview})}

\begin{runningexample}[Running example: Customer support ticket]
A support ticket passes through several stages, Appendix~\ref{appendix:example_workflow_API}.\fullonly{(Figure~\ref{fig:demo_dag})}. 
The first stage, \textit{classify}, uses a lightweight model to extract metadata and determine the handling path. 
The workflow then branches: \textit{route\_ticket} forwards the ticket to an internal team, while \textit{policy\_check} queries company policy using a heavier model and a tool call. 
If needed, the final stage, \textit{reply}, sends the email.
\end{runningexample}







\comp{Stage Mapper}
The stage mapper takes a workflow definition and the available infrastructure as input and produces an execution plan that assigns each stage to a specific model and backend. This mapping reflects the fact that different stages have different computational requirements. \fullonly{A classification stage, for example, is computationally lightweight and latency-sensitive, whereas a code-generation stage that produces a complete function requires a larger, more capable model.}
For each stage, the stage mapper selects the model (e.g., Llama-3, Qwen-2.5, or Mistral), the execution backend (e.g., SGLang on a datacenter GPU or Ollama on an edge device), and the relevant inference parameters (e.g., temperature or structured output mode).

\orla provides two strategies for stage mapping. The first is explicit mapping, where the developer assigns each stage to a specific backend and model. \orla verifies that each stage is assigned to a valid backend and that its inference parameters are compatible with that backend, catching configuration errors before execution begins. The second strategy is dynamic routing through the \texttt{StageMapper} interface. \orla includes a \texttt{OneBitStageMapper} that uses a lightweight LLM call to classify each request as ``simple'' or ``complex'' and routes it to a light or heavy model accordingly. \fullonly{Because the routing call itself runs on the lightweight model, it introduces minimal overhead while enabling automatic request-level routing without developer intervention.}
Developers can also implement custom mappers. 
\fullonly{For example, \orla provides a \texttt{ThresholdStageMapper} that routes requests based on a scoring function (e.g., prompt length) compared to a configurable threshold, requiring no LLM call.}

\noindent\rex\ In the customer support workflow, we use explicit mapping.
The \texttt{classify} stage runs on SGLang with \texttt{Qwen3-4B},
while the \texttt{policy\_check}, \texttt{reply}, and
\texttt{route\_ticket} stages run on a heavier SGLang backend with
\texttt{Qwen3-8B}, as they require stronger reasoning and tool use.



\comp{Workflow Orchestrator}
The workflow orchestrator executes the plan produced by the stage mapper. It consists of two components: stage scheduling and resource and context management.


\noindent\textbf{Scheduling.}
The workflow orchestrator schedules stages according to a configurable policy. Workflow dependencies are enforced before scheduling: a stage is enqueued only when all of its upstream dependencies have completed. \orla organizes ready requests hierarchically: each backend maintains a queue of requests, further partitioned into per-stage sub-queues identified by the stage's globally unique identifier. Scheduling proceeds in two steps. First, a \emph{stage scheduler} selects which stage's queue to serve next. Under the default First Come First Served (FCFS) policy, the scheduler selects the stage whose head request arrived earliest. Alternatively, the scheduler may use priority scheduling, where stages are ordered according to the priority hints provided by the developer. Second, a \emph{request scheduler} selects the next request within the chosen stage queue, also using FCFS by default.

\fullonly{This two-level structure allows \orla to control scheduling at the stage level rather than only at the request level. As a result, policies can prioritize specific workflow stages or differentiate across heterogeneous stage types—capabilities not available in traditional request-level schedulers.}
Developers can easily add custom policies; for example, adding a Shortest Job First (SJF) stage scheduler based on prompt length requires fewer than 20 lines of code.



\noindent\rex\ In the customer support workflow, Appendix~\ref{appendix:mapper_API}, the \textit{classify} stage runs on the light backend using FCFS scheduling. The remaining stages run on the heavy backend using priority scheduling. After \textit{classify} completes, its structured output—such as the ticket category and escalation flag—is used to assign scheduling priorities to subsequent stages. For example, billing or technical tickets are given higher priority than general inquiries. These priorities are propagated through scheduling hints attached to downstream stage requests. This allows the system to prioritize latency-sensitive tickets while less urgent requests wait in the heavy model queue. \fullonly{Figure~\ref{fig:scheduling-code} shows how this policy can be configured using the \orla API.}


\noindent\textbf{Resource and context management.}
The workflow orchestrator also manages execution resources and propagates context between stages as workflows progress.  When a stage completes, its output, such as generated text, structured JSON, or tool results, becomes part of the input context for downstream stages. \orla propagates this information through builder functions attached to each stage. These functions receive the outputs of upstream dependencies and construct the prompt or message context for the next stage before execution begins.

In addition, the orchestrator coordinates execution resources across workflows. It dispatches ready stages to the appropriate backend, maintains backend utilization while other stages wait for tool responses, and allows multiple workflows to execute concurrently across available infrastructure.

\noindent\rex\ In the customer support workflow, Appendix~\ref{appendix:context_mngmnt_API}, the \textit{policy\_check} stage builds its prompt from the output of the upstream \textit{classify} stage.\fullonly{(Figure~\ref{fig:context-code}.)} The \textit{reply} stage's builder similarly consumes both \textit{classify} and \textit{policy\_check}
results, branching its prompt based on the \textit{needs\_escalation} flag from classification.

\comp{Memory Manager}
\fullonly{Agentic workloads interact with KV cache differently from standalone LLM requests.}
Within a workflow, consecutive stages often invoke the same model on the same backend and share large context prefixes: each stage extends the previous context with tool outputs or intermediate results. Preserving the KV cache across these stages can therefore avoid redundant prefilling and significantly reduce latency. At the same time, retaining cache entries for stages that are blocked on tool calls or belong to completed workflows wastes scarce GPU memory. Existing LLM serving systems, such as SGLang, manage KV cache using request-level policies such as LRU eviction, which are unaware of workflow structure. As a result, cache entries may be evicted in the middle of a workflow, forcing expensive re-prefill when the next stage executes.

The memory manager addresses this mismatch by managing the KV cache at the workflow level rather than at the level of individual inference requests. It coordinates cache preservation and reclamation using signals from the workflow orchestrator, such as stage transitions and workflow completion, to determine whether the cached state is likely to be reused by subsequent stages or can be reclaimed to free memory for other workflows. The memory manager supports configurable cache policies. Developers may optionally provide stage-level hints (e.g., \textit{preserve} or \textit{flush}); otherwise, \orla applies its default policy and falls back to the backend’s eviction mechanism when needed.


We implement three policies in the initial release of Orla.
\fullonly{Note that \orla currently only implements these policies for the SGLang LLM backend, as vLLM does not currently ship with a Cache API~\cite{vllm_no_cache_api}.}
\shortonly{\textit{Preserve on small increments} keeps cache when consecutive stages on the same backend extend the context slightly. 
\textit{Flush at workflow boundaries} reclaims cache when workflows complete or switch models/backends. 
\textit{Flush under pressure} evicts cache from idle workflows when memory utilization becomes high.}

\fullonly{\noindent\textbf{Preserve on small increments.} When a new stage begins on the same LLM backend and model as the previous stage, and the increase in tokens between the previous stage and the new stage is below a configurable threshold $\tau$, the Memory Manager instructs the backend to preserve the existing KV cache. Developers can override the threshold per stage, allowing stages that append substantial tool output to benefit from preservation when appropriate. The policy returns no-op when 1) the backend or model changes, 2) there is no preceding stage, or 3) when the token increase exceeds the threshold $\tau$. The intuition here is that when a stage advances the shared context by a small number of tokens, it is beneficial to preserve the KV cache for workflows where consecutive stages on the same model and backend share a large common prefix, as only the new tokens require prefill.\medskip

\noindent\textbf{Flush at workflow boundaries.} The Memory Manager proactively flushes cache when a workflow completes or when transitioning between independent workflows. This frees memory for incoming workflows and prevents stale cache from accumulating. Unlike backend-level LRU eviction, which may flush cache belonging to an active workflow, this policy ensures eviction aligns with logical workflow boundaries. Note that when a new stage begins on a different backend or model from the preceding stage, the policy flushes the \textit{previous} backend's cache for that workflow, since the new backend cannot reuse it. This ensures that the cache is reclaimed at the earliest point where it is no longer useful.\medskip
    
\noindent\textbf{Flush under pressure.} This policy runs on a separate path from the transition-based policy chain. A background pressure monitor periodically polls backends for KV cache utilization. When utilization exceeds a configurable threshold $\tau^{\prime}$, the Memory Manager identifies \textit{idle} workflows, i.e, workflows with preserved cache but no in-flight requests on that backend. The Memory Manager flushes the oldest idle workflow's cache. This greedy approach frees memory quickly while avoiding eviction of cache belonging to active workflows. The intuition here is that when memory pressure is high, the backend's LRU eviction may flush cache at arbitrary points, potentially mid-workflow. The Memory Manager prevents this by proactively flushing cache at stage or workflow boundaries before the backend is forced to evict.}



\noindent\rex\ In the customer support workflow, Appendix~\ref{appendix:memory_API}, \textit{classify} runs on a lightweight backend while later stages run on a heavier backend. The memory manager flushes cache when switching backends, but preserves it when consecutive stages on the same backend share most of their context, avoiding redundant prefilling. All remaining cache is reclaimed when the workflow completes.

\section{Prototype and Demonstration}

\orla is open-source\footnote{\url{github.com/dorcha-inc/orla}} and is demonstrated through the customer support workflow used as the running example.\footnote{Demo video: \href{https://youtu.be/3AbRlYVIRxI}{YouTube demo}}




The live demo presents \orla through a sequence of short examples using a customer support workflow. Each example highlights a different capability of the system and modifies only a few lines of configuration, illustrating how developers can quickly experiment with agentic system behavior using \orla.

Demo users interact with \orla through a live terminal interface. They begin by selecting or typing a customer support ticket. A pre-loaded example ticket is provided, but users may also submit free-form tickets to observe how the system classifies and routes new inputs. Once a ticket is submitted, the terminal displays the workflow DAG and begins execution.

\noindent\textbf{Example 1: Stage mapping.}
The workflow begins with the \textit{classify} stage running on a lightweight model. The terminal prints the structured JSON classification, including the ticket category, product, issue, customer request, and escalation flag. This stage completes quickly, illustrating how \orla maps lightweight tasks to smaller models while reserving heavier models for later stages.

\noindent\textbf{Example 2: Scheduling and parallel execution.}
After classification completes, the DAG executor dispatches the \textit{policy\_check} and \textit{route\_ticket} stages concurrently on a shared heavy backend. Tool calls from both stages interleave in the terminal output. For example, \textit{policy\_check} invokes a \textit{read\_policy\_yaml} tool to retrieve company policy, while \textit{route\_ticket} calls \textit{read\_team\_descriptions} to identify the appropriate internal team. In this example, scheduling uses a priority policy derived from the classification output.

\noindent\textbf{Example 3: Context propagation and cache management.}
The classification output propagates to downstream stages through prompt builders, allowing the workflow to follow different execution paths. For non-escalated tickets, \textit{reply} sends a full resolution email using a \textit{send\_email} tool; for escalated tickets, it sends an acknowledgment while routing the issue to a human team. When stages execute on the same backend with small context changes, the memory manager preserves the KV cache; when workflows complete or switch backends, the cache is flushed.

\vspace{-4pt}

\paragraph{\textbf{Live Artifact}.}
The live demo artifact is the open-source release of \orla,\footnote{\url{https://orlaserver.github.io}} accompanied by a step-by-step tutorial. The tutorial provides two reproduction paths. The GPU path runs the workflow on SGLang or vLLM backends serving \texttt{Qwen3-4B} and \texttt{Qwen3-8B}, suitable for users with GPU access. The laptop path runs the same workflow on Ollama backends with lightweight \texttt{Qwen3} models (0.6B and 1.7B), requiring only Docker and a CPU and running locally on a typical laptop.

In both cases, setup consists of a single \texttt{docker-compose} command that launches the backends and the \orla server. Switching between backends requires changing only a single environment variable. The workflow definition, scheduling policies, memory configuration, and tools remain identical across backends, illustrating \orla's plug-and-play backend design.

\section{Evaluation}

We evaluate \orla on two datasets. 
First, we use the SWE-bench Lite benchmark~\cite{jimenez2023swe} consisting of 2,294 requests; the setup is described in Appendix~\ref{appendix:eval}.  
For each instance, we extract the files modified by the gold patch, read them from the repository at the base commit, and construct a prompt asking the model to generate a unified diff that fixes the bug. Second, to evaluate \orla's agent-granularity cache management, we use the DAG-MATH-Formatted-CoT dataset~\cite{llmsdag} with 2,894 requests. We evaluate the first five problems (sorted by problem ID), each executed as a multi-stage workflow. We compare two KV cache strategies: (i) \emph{flush per request}, where each stage sets \texttt{CachePolicy: flush}, evicting the cache after every LLM call, and (ii) \emph{flush per workflow}, where \orla's default policy preserves cache across stages and flushes it only when the workflow completes.

Baseline: All requests run on \texttt{Qwen3-8B} with vLLM.
For mapping experiment, we used \texttt{OneBitStageMapper} that routes simple requests to \texttt{Qwen3-4B} and complex ones to \texttt{Qwen3-8B} on two vLLM backends serving. For cache management experiment, we used an SGLang backend with \texttt{Qwen3-8B}.



\paragraph{\textbf{Results}}

\begin{figure}[t]
\centering

\begin{subfigure}{0.24\linewidth}
  \centering
  \includegraphics[width=\linewidth]{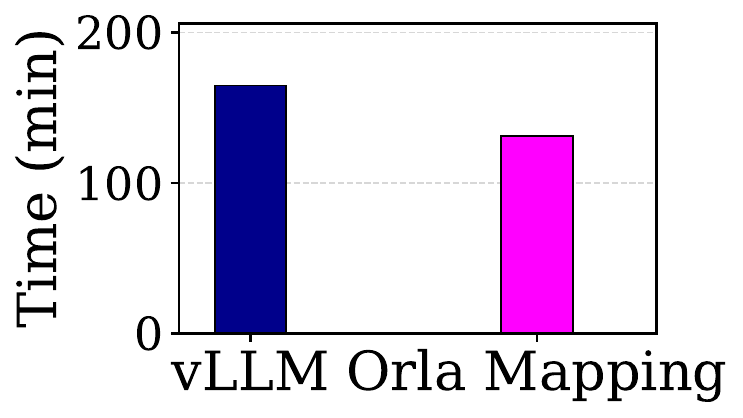}
  \caption{Duration}
  \label{fig:duration_bar}
\end{subfigure}
\hfill
\begin{subfigure}{0.24\linewidth}
  \centering
  \includegraphics[width=\linewidth]{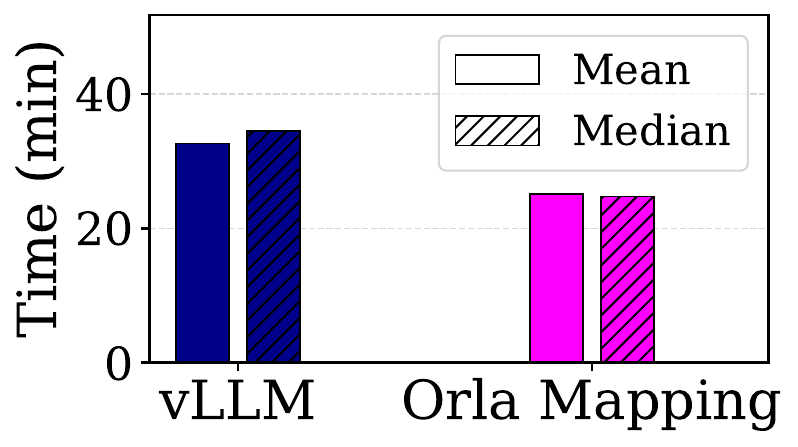}
  \caption{Completion}
  \label{fig:avg_completion}
\end{subfigure}
\hfill
\begin{subfigure}{0.24\linewidth}
  \centering
  \includegraphics[width=\linewidth]{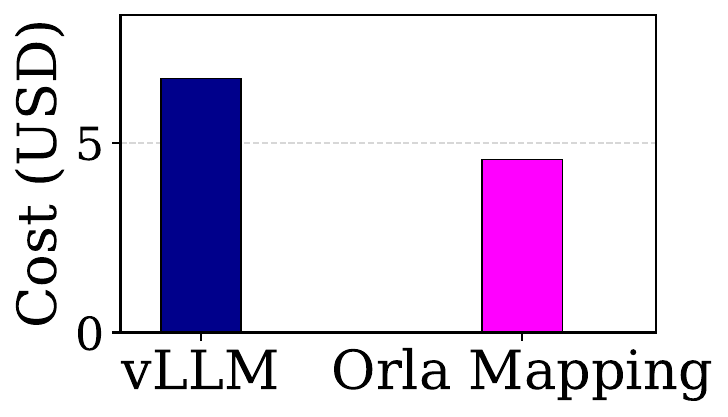}
  \caption{Cost}
  \label{fig:cost_bar}
\end{subfigure}

\caption{(a) Wall-clock time (b) Cost (c) Completion time.}
\label{fig:performance_bars}
\end{figure}

Stage mapping improves both latency and cost. 
Out of 2,294 requests, 956 are routed to the lightweight 
\texttt{Qwen3-4B} model and the rest execute on 
\texttt{Qwen3-8B}. This heterogeneous execution reduces wall-clock time by 38\% 
(Figure~\ref{fig:duration_bar}) and mean completion time by 60\% 
(Figure~\ref{fig:avg_completion}) compared to the single-model baseline. 
Because the lightweight model is cheaper per token in cloud deployments, stage mapping also reduces estimated inference cost by 35\% 
(Figure~\ref{fig:cost_bar}).

\begin{figure}
    \centering
    \includegraphics[width=0.4\linewidth]{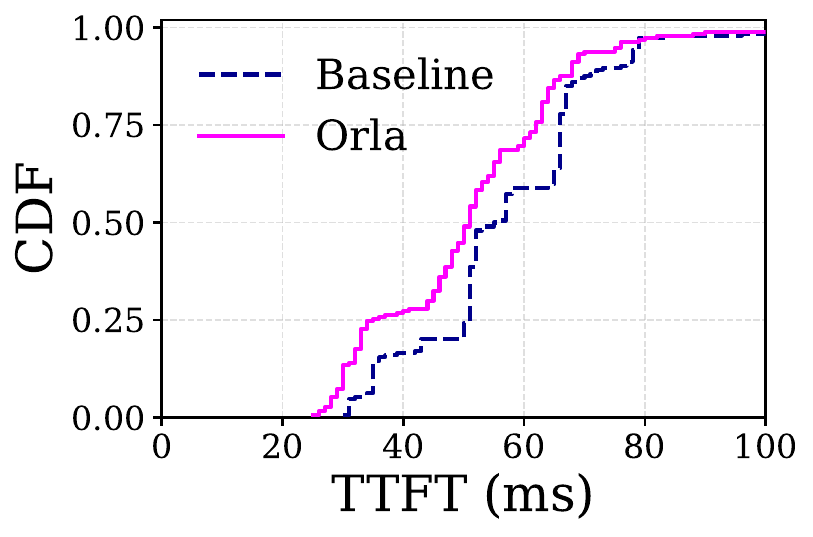}
    \caption{Cache management (request vs. workflow flush).}
    \label{fig:ttft_cdf}
\end{figure}

Preserving KV cache across stages further reduces latency. 
Figure~\ref{fig:ttft_cdf} shows the CDF of time-to-first-token (TTFT) for the two cache strategies. 
Compared to flush-per-request, Orla's workflow-level cache management 
reduces TTFT and improves the entire latency distribution by enabling 
KV cache reuse across stages.

\section{Conclusion}

We introduced \orla, a library for constructing and running LLM-based agentic systems. \orla separates request execution from workflow-level policy through three components: a \emph{stage mapper}, a \emph{workflow orchestrator}, and a \emph{memory manager}, which manages inference state such as the KV cache across workflow boundaries.

\orla treats all backends as plug-and-play modules accessed through a simple uniform interface, making it suitable both as a serving library for agentic systems and as a research testbed. Our evaluation shows that stage mapping enables efficient heterogeneous execution, reducing both latency and inference cost without modifying the underlying LLM serving engine.

\newpage
\bibliographystyle{ACM-Reference-Format}

\begin{thebibliography}{18}


\ifx \showCODEN    \undefined \def \showCODEN     #1{\unskip}     \fi
\ifx \showISBNx    \undefined \def \showISBNx     #1{\unskip}     \fi
\ifx \showISBNxiii \undefined \def \showISBNxiii  #1{\unskip}     \fi
\ifx \showISSN     \undefined \def \showISSN      #1{\unskip}     \fi
\ifx \showLCCN     \undefined \def \showLCCN      #1{\unskip}     \fi
\ifx \shownote     \undefined \def \shownote      #1{#1}          \fi
\ifx \showarticletitle \undefined \def \showarticletitle #1{#1}   \fi
\ifx \showURL      \undefined \def \showURL       {\relax}        \fi
\providecommand\bibfield[2]{#2}
\providecommand\bibinfo[2]{#2}
\providecommand\natexlab[1]{#1}
\providecommand\showeprint[2][]{arXiv:#2}

\bibitem[{Bloomberg}(2026)]%
        {Bloomberg2026ASKB}
\bibfield{author}{\bibinfo{person}{{Bloomberg}}.}
  \bibinfo{year}{2026}\natexlab{}.
\newblock \bibinfo{title}{Meet {ASKB}: {Bloomberg} Introduces Agentic {AI} to
  the {Bloomberg} Terminal}.
\newblock
\urldef\tempurl%
\url{https://www.bloomberg.com/company/stories/meet-askb-bloomberg-introduces-agentic-ai-to-the-bloomberg-terminal/}
\showURL{%
\tempurl}
\newblock
\shownote{Accessed: 2026-03-06}.


\bibitem[Gao et~al\mbox{.}(2025)]%
        {gao2025survey}
\bibfield{author}{\bibinfo{person}{Huan-ang Gao}, \bibinfo{person}{Jiayi Geng},
  \bibinfo{person}{Wenyue Hua}, \bibinfo{person}{Mengkang Hu},
  \bibinfo{person}{Xinzhe Juan}, \bibinfo{person}{Hongzhang Liu},
  \bibinfo{person}{Shilong Liu}, \bibinfo{person}{Jiahao Qiu},
  \bibinfo{person}{Xuan Qi}, \bibinfo{person}{Yiran Wu}, {et~al\mbox{.}}}
  \bibinfo{year}{2025}\natexlab{}.
\newblock \showarticletitle{A survey of self-evolving agents: On path to
  artificial super intelligence}.
\newblock \bibinfo{journal}{\emph{arXiv preprint arXiv:2507.21046}}
  \bibinfo{volume}{1} (\bibinfo{year}{2025}).
\newblock


\bibitem[Gottweis et~al\mbox{.}(2025)]%
        {gottweis2025towards}
\bibfield{author}{\bibinfo{person}{Juraj Gottweis}, \bibinfo{person}{Wei-Hung
  Weng}, \bibinfo{person}{Alexander Daryin}, \bibinfo{person}{Tao Tu},
  \bibinfo{person}{Anil Palepu}, \bibinfo{person}{Petar Sirkovic},
  \bibinfo{person}{Artiom Myaskovsky}, \bibinfo{person}{Felix Weissenberger},
  \bibinfo{person}{Keran Rong}, \bibinfo{person}{Ryutaro Tanno},
  {et~al\mbox{.}}} \bibinfo{year}{2025}\natexlab{}.
\newblock \showarticletitle{Towards an AI co-scientist}.
\newblock \bibinfo{journal}{\emph{arXiv preprint arXiv:2502.18864}}
  (\bibinfo{year}{2025}).
\newblock


\bibitem[Jimenez et~al\mbox{.}(2023)]%
        {jimenez2023swe}
\bibfield{author}{\bibinfo{person}{Carlos~E Jimenez}, \bibinfo{person}{John
  Yang}, \bibinfo{person}{Alexander Wettig}, \bibinfo{person}{Shunyu Yao},
  \bibinfo{person}{Kexin Pei}, \bibinfo{person}{Ofir Press}, {and}
  \bibinfo{person}{Karthik Narasimhan}.} \bibinfo{year}{2023}\natexlab{}.
\newblock \showarticletitle{Swe-bench: Can language models resolve real-world
  github issues?}
\newblock \bibinfo{journal}{\emph{arXiv preprint arXiv:2310.06770}}
  (\bibinfo{year}{2023}).
\newblock


\bibitem[Kang et~al\mbox{.}(2026)]%
        {kang2026thunderagent}
\bibfield{author}{\bibinfo{person}{Hao Kang}, \bibinfo{person}{Ziyang Li},
  \bibinfo{person}{Xinyu Yang}, \bibinfo{person}{Weili Xu},
  \bibinfo{person}{Yinfang Chen}, \bibinfo{person}{Junxiong Wang},
  \bibinfo{person}{Beidi Chen}, \bibinfo{person}{Tushar Krishna},
  \bibinfo{person}{Chenfeng Xu}, {and} \bibinfo{person}{Simran Arora}.}
  \bibinfo{year}{2026}\natexlab{}.
\newblock \bibinfo{title}{ThunderAgent: A Simple, Fast and Program-Aware
  Agentic Inference System}.
\newblock
\showeprint[arxiv]{2602.13692}~[cs.OS]
\urldef\tempurl%
\url{https://arxiv.org/abs/2602.13692}
\showURL{%
\tempurl}


\bibitem[Khattab et~al\mbox{.}(2023)]%
        {khattab2023dspy}
\bibfield{author}{\bibinfo{person}{Omar Khattab}, \bibinfo{person}{Arnav
  Singhvi}, \bibinfo{person}{Paridhi Maheshwari}, \bibinfo{person}{Zhiyuan
  Zhang}, \bibinfo{person}{Keshav Santhanam}, \bibinfo{person}{Sri
  Vardhamanan}, \bibinfo{person}{Saiful Haq}, \bibinfo{person}{Ashutosh
  Sharma}, \bibinfo{person}{Thomas~T Joshi}, \bibinfo{person}{Hanna Moazam},
  {et~al\mbox{.}}} \bibinfo{year}{2023}\natexlab{}.
\newblock \showarticletitle{Dspy: Compiling declarative language model calls
  into self-improving pipelines}.
\newblock \bibinfo{journal}{\emph{arXiv preprint arXiv:2310.03714}}
  (\bibinfo{year}{2023}).
\newblock


\bibitem[Kwon et~al\mbox{.}(2023)]%
        {kwon2023efficient}
\bibfield{author}{\bibinfo{person}{Woosuk Kwon}, \bibinfo{person}{Zhuohan Li},
  \bibinfo{person}{Siyuan Zhuang}, \bibinfo{person}{Ying Sheng},
  \bibinfo{person}{Lianmin Zheng}, \bibinfo{person}{Cody~Hao Yu},
  \bibinfo{person}{Joseph Gonzalez}, \bibinfo{person}{Hao Zhang}, {and}
  \bibinfo{person}{Ion Stoica}.} \bibinfo{year}{2023}\natexlab{}.
\newblock \showarticletitle{Efficient memory management for large language
  model serving with pagedattention}. In \bibinfo{booktitle}{\emph{Proceedings
  of the 29th symposium on operating systems principles}}.
  \bibinfo{pages}{611--626}.
\newblock


\bibitem[LangChain(2026)]%
        {langgraph}
\bibfield{author}{\bibinfo{person}{LangChain}.}
  \bibinfo{year}{2026}\natexlab{}.
\newblock \bibinfo{title}{LangGraph}.
\newblock \bibinfo{howpublished}{\url{https://www.langchain.com/langgraph}}.
\newblock


\bibitem[Mitzenmacher and Shahout(2025)]%
        {mitzenmacher2025queueing}
\bibfield{author}{\bibinfo{person}{Michael Mitzenmacher} {and}
  \bibinfo{person}{Rana Shahout}.} \bibinfo{year}{2025}\natexlab{}.
\newblock \showarticletitle{Queueing, predictions, and large language models:
  Challenges and open problems}.
\newblock \bibinfo{journal}{\emph{Stochastic Systems}} \bibinfo{volume}{15},
  \bibinfo{number}{3} (\bibinfo{year}{2025}), \bibinfo{pages}{195--219}.
\newblock


\bibitem[Patel et~al\mbox{.}(2024)]%
        {patel2024splitwise}
\bibfield{author}{\bibinfo{person}{Pratyush Patel}, \bibinfo{person}{Esha
  Choukse}, \bibinfo{person}{Chaojie Zhang}, \bibinfo{person}{Aashaka Shah},
  \bibinfo{person}{{\'I}{\~n}igo Goiri}, \bibinfo{person}{Saeed Maleki}, {and}
  \bibinfo{person}{Ricardo Bianchini}.} \bibinfo{year}{2024}\natexlab{}.
\newblock \showarticletitle{Splitwise: Efficient generative llm inference using
  phase splitting}. In \bibinfo{booktitle}{\emph{2024 ACM/IEEE 51st Annual
  International Symposium on Computer Architecture (ISCA)}}. IEEE,
  \bibinfo{pages}{118--132}.
\newblock


\bibitem[Shahout et~al\mbox{.}(2024a)]%
        {shahout2024fast}
\bibfield{author}{\bibinfo{person}{Rana Shahout}, \bibinfo{person}{Cong Liang},
  \bibinfo{person}{Shiji Xin}, \bibinfo{person}{Qianru Lao},
  \bibinfo{person}{Yong Cui}, \bibinfo{person}{Minlan Yu}, {and}
  \bibinfo{person}{Michael Mitzenmacher}.} \bibinfo{year}{2024}\natexlab{a}.
\newblock \showarticletitle{Fast inference for augmented large language
  models}.
\newblock \bibinfo{journal}{\emph{arXiv preprint arXiv:2410.18248}}
  (\bibinfo{year}{2024}).
\newblock


\bibitem[Shahout et~al\mbox{.}(2024b)]%
        {shahout2024don}
\bibfield{author}{\bibinfo{person}{Rana Shahout}, \bibinfo{person}{Eran
  Malach}, \bibinfo{person}{Chunwei Liu}, \bibinfo{person}{Weifan Jiang},
  \bibinfo{person}{Minlan Yu}, {and} \bibinfo{person}{Michael Mitzenmacher}.}
  \bibinfo{year}{2024}\natexlab{b}.
\newblock \showarticletitle{Don't Stop Me Now: Embedding Based Scheduling for
  LLMs}.
\newblock \bibinfo{journal}{\emph{arXiv preprint arXiv:2410.01035}}
  (\bibinfo{year}{2024}).
\newblock


\bibitem[Wu et~al\mbox{.}(2024)]%
        {wu2024autogen}
\bibfield{author}{\bibinfo{person}{Qingyun Wu}, \bibinfo{person}{Gagan Bansal},
  \bibinfo{person}{Jieyu Zhang}, \bibinfo{person}{Yiran Wu},
  \bibinfo{person}{Beibin Li}, \bibinfo{person}{Erkang~(Eric) Zhu},
  \bibinfo{person}{Li Jiang}, \bibinfo{person}{Xiaoyun Zhang},
  \bibinfo{person}{Shaokun Zhang}, \bibinfo{person}{Ahmed Awadallah},
  \bibinfo{person}{Ryen~W. White}, \bibinfo{person}{Doug Burger}, {and}
  \bibinfo{person}{Chi Wang}.} \bibinfo{year}{2024}\natexlab{}.
\newblock \showarticletitle{AutoGen: Enabling Next-Gen LLM Applications via
  Multi-Agent Conversation}. In \bibinfo{booktitle}{\emph{COLM 2024}}.
\newblock
\urldef\tempurl%
\url{https://www.microsoft.com/en-us/research/publication/autogen-enabling-next-gen-llm-applications-via-multi-agent-conversation-framework/}
\showURL{%
\tempurl}


\bibitem[Wu et~al\mbox{.}(2023)]%
        {wu2023bloomberggpt}
\bibfield{author}{\bibinfo{person}{Shijie Wu}, \bibinfo{person}{Ozan Irsoy},
  \bibinfo{person}{Steven Lu}, \bibinfo{person}{Vadim Dabravolski},
  \bibinfo{person}{Mark Dredze}, \bibinfo{person}{Sebastian Gehrmann},
  \bibinfo{person}{Prabhanjan Kambadur}, \bibinfo{person}{David Rosenberg},
  {and} \bibinfo{person}{Gideon Mann}.} \bibinfo{year}{2023}\natexlab{}.
\newblock \showarticletitle{Bloomberggpt: A large language model for finance}.
\newblock \bibinfo{journal}{\emph{arXiv preprint arXiv:2303.17564}}
  (\bibinfo{year}{2023}).
\newblock


\bibitem[Xia et~al\mbox{.}(2025)]%
        {xia2025live}
\bibfield{author}{\bibinfo{person}{Chunqiu~Steven Xia}, \bibinfo{person}{Zhe
  Wang}, \bibinfo{person}{Yan Yang}, \bibinfo{person}{Yuxiang Wei}, {and}
  \bibinfo{person}{Lingming Zhang}.} \bibinfo{year}{2025}\natexlab{}.
\newblock \showarticletitle{Live-SWE-agent: Can Software Engineering Agents
  Self-Evolve on the Fly?}
\newblock \bibinfo{journal}{\emph{arXiv preprint arXiv:2511.13646}}
  (\bibinfo{year}{2025}).
\newblock


\bibitem[Yu et~al\mbox{.}(2022)]%
        {yu2022orca}
\bibfield{author}{\bibinfo{person}{Gyeong-In Yu}, \bibinfo{person}{Joo~Seong
  Jeong}, \bibinfo{person}{Geon-Woo Kim}, \bibinfo{person}{Soojeong Kim}, {and}
  \bibinfo{person}{Byung-Gon Chun}.} \bibinfo{year}{2022}\natexlab{}.
\newblock \showarticletitle{Orca: A distributed serving system for
  $\{$Transformer-Based$\}$ generative models}. In
  \bibinfo{booktitle}{\emph{16th USENIX symposium on operating systems design
  and implementation (OSDI 22)}}. \bibinfo{pages}{521--538}.
\newblock


\bibitem[Zhang et~al\mbox{.}(2025)]%
        {llmsdag}
\bibfield{author}{\bibinfo{person}{Yuanhe Zhang}, \bibinfo{person}{Ilja
  Kuzborskij}, \bibinfo{person}{Jason~D. Lee}, \bibinfo{person}{Chenlei Leng},
  {and} \bibinfo{person}{Fanghui Liu}.} \bibinfo{year}{2025}\natexlab{}.
\newblock \showarticletitle{DAG-Math: Graph-Guided Mathematical Reasoning in
  LLMs}.
\newblock \bibinfo{journal}{\emph{arXiv preprint arXiv:2510.19842}}
  (\bibinfo{year}{2025}).
\newblock


\bibitem[Zheng et~al\mbox{.}(2024)]%
        {zheng2024sglang}
\bibfield{author}{\bibinfo{person}{Lianmin Zheng}, \bibinfo{person}{Liangsheng
  Yin}, \bibinfo{person}{Zhiqiang Xie}, \bibinfo{person}{Chuyue~Livia Sun},
  \bibinfo{person}{Jeff Huang}, \bibinfo{person}{Cody~Hao Yu},
  \bibinfo{person}{Shiyi Cao}, \bibinfo{person}{Christos Kozyrakis},
  \bibinfo{person}{Ion Stoica}, \bibinfo{person}{Joseph~E Gonzalez},
  {et~al\mbox{.}}} \bibinfo{year}{2024}\natexlab{}.
\newblock \showarticletitle{Sglang: Efficient execution of structured language
  model programs}.
\newblock \bibinfo{journal}{\emph{Advances in neural information processing
  systems}}  \bibinfo{volume}{37} (\bibinfo{year}{2024}),
  \bibinfo{pages}{62557--62583}.
\newblock


\end{thebibliography}

\newpage
\appendix
\section{Running Example Workflow and API}
\label{appendix:example_workflow_API}

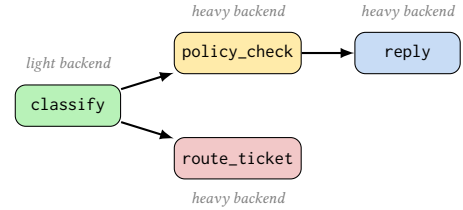
\begin{figure}[H]
\centering
\begin{tikzpicture}[
    node distance=0.3cm and 0.7cm,
    stage/.style={rectangle, draw, rounded corners,
      minimum width=1.4cm, minimum height=0.55cm,
      font=\footnotesize\ttfamily},
    lbl/.style={font=\scriptsize\itshape, text=gray},
    >=latex
]
\node[stage, fill=stagegreen] (classify) {classify};
\node[stage, right=of classify, yshift=0.7cm, fill=stageyellow] (policy) {policy\_check};
\node[stage, right=of policy, fill=stageblue] (reply) {reply};
\node[stage, right=of classify, yshift=-0.7cm, fill=stagepink] (route) {route\_ticket};
\node[lbl, above=0.05cm of classify] {light backend};
\node[lbl, above=0.05cm of policy] {heavy backend};
\node[lbl, above=0.05cm of reply] {heavy backend};
\node[lbl, below=0.05cm of route] {heavy backend};
\draw[->,thick] (classify) -- (policy);
\draw[->,thick] (policy) -- (reply);
\draw[->,thick] (classify) -- (route);
\end{tikzpicture}
\caption{The workflow of our running example, a customer support workflow.}
\label{fig:demo_dag}
\end{figure}

\begin{figure}[H]
\centering
\begin{tcolorbox}[
    title={Multi-Agent Workflow Definition in \orla},
    fonttitle=\bfseries\small,
    coltitle=black,
    colbacktitle=gray!20,
    colback=gray!2,
    colframe=gray!75,
    arc=1mm,
    left=5pt, right=5pt, top=5pt, bottom=5pt,
    toptitle=2pt, bottomtitle=2pt
]

\begin{codeblock}
// 1. Initialize Backend Clients
client := orla.NewOrlaClient("http://localhost:8081")
light  := orla.NewSGLangBackend("Qwen3-4B", lightURL)
heavy  := orla.NewSGLangBackend("Qwen3-8B", heavyURL)

// 2. Define Workflow Stages
classify := orla.NewStage("classify", light)
classify.SetResponseFormat(
    orla.NewStructuredOutputRequest("ticket_classify", 
                                     classifySchema)
)

policy := orla.NewStage("policy_check", heavy)
policy.SetExecutionMode(orla.ExecutionModeAgentLoop)
policy.AddTool(readPolicyTool)

reply := orla.NewStage("reply", heavy)
reply.SetExecutionMode(orla.ExecutionModeAgentLoop)
reply.AddTool(sendEmailTool)

route := orla.NewStage("route_ticket", heavy)
route.SetExecutionMode(orla.ExecutionModeAgentLoop)

// 3. Assemble and Execute DAG
wf := orla.NewWorkflow(client)
wf.AddStage(classify, policy, reply, route)

wf.AddDependency(policy.ID, classify.ID) // Classify -> Policy
wf.AddDependency(reply.ID, policy.ID)   // Policy   -> Reply
wf.AddDependency(route.ID, classify.ID) // Classify -> Route

results, _ := wf.Execute(ctx)
\end{codeblock}

\end{tcolorbox}
\caption{The \orla API allows users to define complex agentic workflows using a DAG-based approach, as shown in this customer support example.}
\label{fig:workflow-code}
\end{figure}

\section{API for \orla Components}

\subsection{State mapper API}
\label{appendix:mapper_API}

\begin{figure}[H]
\begin{tcolorbox}[
    title={Configuring Backends using Orla's API},
    fonttitle=\bfseries\small,
    coltitle=black,
    colbacktitle=gray!20,
    colback=gray!2,
    colframe=gray!75,
    arc=1mm,
    left=5pt, right=5pt, top=5pt, bottom=5pt,
    toptitle=2pt, bottomtitle=2pt
]
\begin{codeblock}
// Production: SGLang on datacenter GPU
light := orla.NewSGLangBackend("Qwen3-4B", gpuURL)

// Testing: Ollama on laptop CPU
light := orla.NewOllamaBackend("qwen3:0.6b", ollamaURL)

// Research and Simulation: latency model, no LLM
light := orla.NewSimulatedBackend("sim-light", simURL)
\end{codeblock}
\end{tcolorbox}
\end{figure}

\subsection{Scheduling API}
\label{appendix:scheduling_API}

\begin{figure}[H]
\begin{tcolorbox}[
    title={Specifying a Scheduling Policy in \orla},
    fonttitle=\bfseries\small,
    coltitle=black,
    colbacktitle=gray!20,
    colback=gray!2,
    colframe=gray!75,
    arc=1mm,
    left=5pt, right=5pt, top=5pt, bottom=5pt,
    toptitle=2pt, bottomtitle=2pt
]

\begin{codeblock}
// Priority scheduling across stage queues
classify.SetSchedulingPolicy("priority")
policy.SetSchedulingPolicy("priority")

// FIFO within each stage's queue
classify.SetRequestSchedulingPolicy("fifo")

.........

// Dynamic priority, updated at runtime based on 
// the classify stage's output
priority := 5
if classifyData.Category == "billing" 
   || classifyData.Category == "technical" {
    priority = 8
}

replyStage.SetSchedulingHints(
  &orla.SchedulingHints{Priority: &priority})
})
\end{codeblock}

\end{tcolorbox}
\caption{An illustration of \orla's two-level scheduling policy in our running example. Stage scheduling policy selects which stage sub-queue to dequeue from, while request scheduling policy orders requests within a sub-queue. Priority hints can be set dynamically based on ticket classification.}
\label{fig:scheduling-code}
\end{figure}

\subsection{Context management API}
\label{appendix:context_mngmnt_API}

\begin{figure}[H]
\begin{tcolorbox}[
    title={Context passing via Orla's API},
    fonttitle=\bfseries\small,
    coltitle=black,
    colbacktitle=gray!20,
    colback=gray!2,
    colframe=gray!75,
    arc=1mm,
    left=5pt, right=5pt, top=5pt, bottom=5pt,
    toptitle=2pt, bottomtitle=2pt
]

\begin{codeblock}
policy.SetPromptBuilder(
  func(upstream map[string]*orla.StageResult) (string, error) {
    classifyResult := upstream[classify.ID]
    return fmt.Sprintf(
      "Review this classification against company policy.\n"+
        "Classification: %s\nOriginal ticket: %s",
      classifyResult.Response.Content, ticket), nil
  })
\end{codeblock}

\end{tcolorbox}
\caption{The \textit{policy\_check} stage constructs its prompt from the upstream \textit{classify} stage's result.}
\label{fig:context-code}
\end{figure}

\subsection{Memory manager API}
\label{appendix:memory_API}

\begin{figure}[H]
\begin{tcolorbox}[
    title={Memory management via Orla's API},
    fonttitle=\bfseries\small,
    coltitle=black,
    colbacktitle=gray!20,
    colback=gray!2,
    colframe=gray!75,
    arc=1mm,
    left=5pt, right=5pt, top=5pt, bottom=5pt,
    toptitle=2pt, bottomtitle=2pt
]

\begin{codeblock}
// At the workflow level, we use the flush at workflow 
// boundary policy
wf.SetMemoryPolicy(orla.NewFlushAtBoundaryPolicy())

// Force flush when entering policy_check since we 
// know that classify uses light backend only once. 
// Note that this line is not strictly needed, as Orla's 
// memory manager will recognize a backend change automatically.
policyStage.SetCachePolicy("flush")

// Preserve between policy_check and reply since we know 
// that the same backend is used for the shared context
replyStage.SetCachePolicy("preserve")

// Run workflow 1, then some other workflow 2. 
// The memory manager will handle the cache management for us.
wf.Execute(ctx)
wf2.Execute(ctx)

// The memory manager will flush the light backend at 
// the classify->policy_check boundary,
// preserve the cache at the policy_check->reply boundary,
// and flush the cache at the workflow boundary when wf2 starts.
\end{codeblock}

\end{tcolorbox}
\caption{In the customer support workflow, \textit{classify} runs on a lightweight backend while later stages run on a heavier backend. The memory manager flushes cache when switching backends, but preserves it when consecutive stages on the same backend share most of their context, avoiding redundant prefilling. All remaining cache is reclaimed when the workflow completes.}
\label{fig:context-code}
\end{figure}

\section{Evaluation}
\label{appendix:eval}

\subsection{Setup}

All experiments run on a single machine with two AMD EPYC 7313 16-Core Processors (64 cores total, 2 threads per core), 528\,GB of system memory, and one NVIDIA RTX PRO 6000 Blackwell Max-Q Workstation Edition GPU with 96\,GB VRAM, running CUDA 13.0 on Linux kernel 5.15.0. For the stage mapping experiment, we used vLLM to serve both models on the same GPU: the heavy model (Qwen3-8B) is allocated 60\% of GPU memory and the light model (Qwen3-4B) receives 35\%, with a maximum context length of 32{,}768 tokens each.The \orla server runs as a separate container orchestrating inference requests to both backends. Temperature is set to 0 for reproducibility and maximum output length is capped at 4{,}096 tokens. For the cache management experiment, we used a Qwen3-8B model served on SGLang as SGLang supports a KVCache management API. The temperature is still set to 0 and the maximum output length is capped at 256.

\end{document}